\pdfoutput=1
\documentclass[11pt]{article}

\usepackage[preprint]{acl}

\usepackage{times}
\usepackage{latexsym}
\usepackage[T1]{fontenc}
\usepackage[utf8]{inputenc}

\usepackage{microtype}

\usepackage{inconsolata}

\usepackage{graphicx}

\usepackage{hyperref}
\usepackage{url}

\usepackage{graphicx}
\usepackage{subfigure}
\usepackage{wrapfig}

\usepackage{amsmath}
\usepackage{amstext}
\usepackage{amsfonts}
\usepackage{bm}

\usepackage{bbm}
\usepackage{multirow}
\usepackage{booktabs}
\usepackage{array}
\usepackage{caption}
\usepackage{multirow}
\usepackage{booktabs}
\usepackage{array}
\usepackage{caption}
\usepackage{color}
\usepackage{colortbl}
\usepackage{tablefootnote}
\usepackage{adjustbox}

\usepackage{caption}
\usepackage{algorithm}
\usepackage{algpseudocode}

\algnewcommand{\LineComment}[1]{\Statex ~~~~~~\textsc{//}~\textit{#1}}

\usepackage{enumitem}
\setenumerate[1]{itemsep=0pt,partopsep=0pt,parsep=\parskip,topsep=5pt}
\setitemize[1]{itemsep=0pt,partopsep=0pt,parsep=\parskip,topsep=5pt}
\setdescription{itemsep=0pt,partopsep=0pt,parsep=\parskip,topsep=5pt}

\usepackage{soul}

\usepackage{tikz}
\usepackage[edges]{forest}
\definecolor{hidden-draw}{RGB}{64,101,149}
\definecolor{hidden-pink}{RGB}{231,239,250}

\usepackage[mathscr]{euscript}
\usepackage{amssymb}  
\usepackage{pifont}

\usepackage{arydshln}

\usepackage{ulem}

\title{\texorpdfstring{\includegraphics[width=26pt]{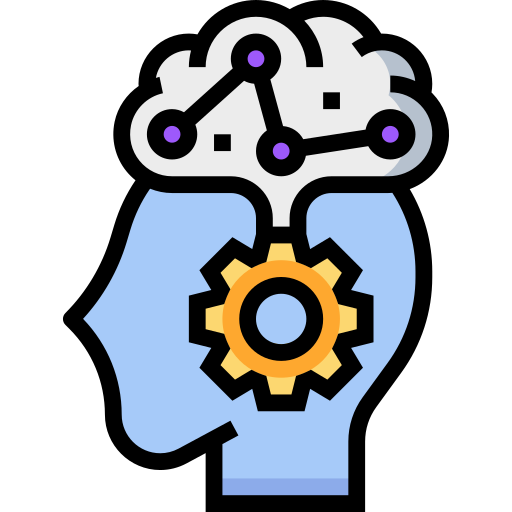}}{}~\textit{Reasoning Beyond Language:} A Comprehensive Survey on \\ Latent Chain-of-Thought Reasoning}

\author{{Xinghao Chen}\textsuperscript{\rm 1,2}\thanks{Equal Contributions.}, {Anhao Zhao}\textsuperscript{\rm 1,2}\footnotemark[1], {Heming Xia}\textsuperscript{\rm 1}, {Xuan Lu}\textsuperscript{\rm 2}, {\textbf{Hanlin Wang}}\textsuperscript{\rm 1}, \\
{\textbf{Yanjun Chen}}\textsuperscript{\rm 1,2}, {\textbf{Wei Zhang}}\textsuperscript{\rm 2}, {\textbf{Jian Wang}}\textsuperscript{\rm 1}\thanks{Corresponding Authors.}, {\textbf{Wenjie Li}}\textsuperscript{\rm 1}, {\textbf{Xiaoyu Shen}}\textsuperscript{\rm 2}\footnotemark[2]\\
  \textsuperscript{\rm 1}Department of Computing, The Hong Kong Polytechnic University \\
  \textsuperscript{\rm 2}Zhejiang Key Laboratory of Industrial Intelligence and Digital Twin, EIT, Ningbo\\
  {\tt xing-hao.chen@connect.polyu.hk ~ xyshen@eitech.edu.cn}
}

\begin{document}
\maketitle
\begin{abstract}
Large Language Models (LLMs) have shown impressive performance on complex tasks through Chain-of-Thought (CoT) reasoning. 
However, conventional CoT relies on explicitly verbalized intermediate steps, which constrains its broader applicability, particularly in abstract reasoning tasks beyond language.
To address this, there has been growing research interest in \textit{latent CoT reasoning}, where the reasoning process is embedded within latent spaces. 
By decoupling reasoning from explicit language generation, latent CoT offers the promise of richer cognitive representations and facilitates more flexible, faster inference.
This paper aims to present a comprehensive overview of this emerging paradigm and establish a systematic taxonomy. 
We analyze recent advances in methods, categorizing them from token-wise horizontal approaches to layer-wise vertical strategies.
We then provide in-depth discussions of these methods, highlighting their design principles, applications, and remaining challenges. We hope that our survey provides a structured foundation for advancing this promising direction in LLM reasoning.
\footnote{The list of relevant papers will be regularly updated at \url{https://github.com/EIT-NLP/Awesome-Latent-CoT}.}

\end{abstract}
\section{Introduction}
\label{sec:intro}

\begin{quote}
    \raggedright \textit{Whereof one cannot speak, thereof one must be silent.} 
    \hfill --- Ludwig Wittgenstein
\end{quote}

The Chain-of-Thought (CoT) paradigm~\citep{wei2022chain}, which encourages large language models (LLMs) to articulate their reasoning step by step in natural language, has substantially advanced both interpretability and performance on many tasks~\citep{guo2025deepseek, openai_learning_to_reason, qwen3_Think_Deeper_Act_Faster, chen2025reasoningerasurveylong}. By verbalizing intermediate steps, CoT provides not only a transparent view of the model’s decision process but also a mechanism for human intervention.

\begin{figure}[t]
\centering
    \includegraphics[width=1.0\columnwidth]{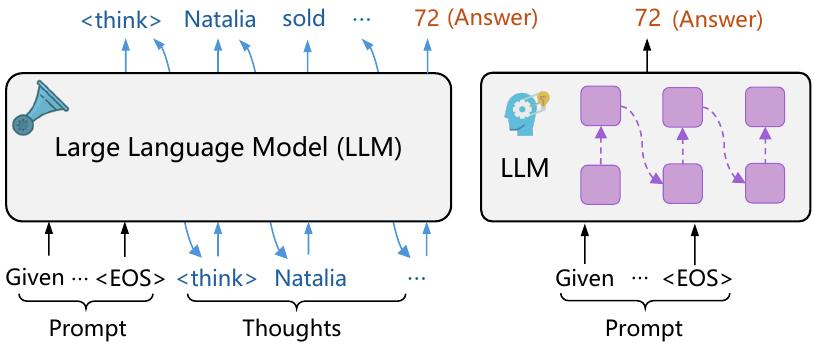}
    \vspace{-16pt}
    \caption{Explicit CoT (left) generates reasoning steps with natural language, while latent CoT (right) allows the model to reason internally in latent spaces. 
    }
    \vspace{-16pt}
    \label{fig:intro}
\end{figure}

\definecolor{maingray}{RGB}{230,230,230}
\definecolor{darkgray}{RGB}{100,100,100}
\definecolor{lightpurple}{RGB}{235,230,250}
\definecolor{darkpurple}{RGB}{120,90,160}
\definecolor{lightblue}{RGB}{225,240,255}
\definecolor{darkblue}{RGB}{70,120,200}
\definecolor{lightpink}{RGB}{255,235,240}
\definecolor{darkpink}{RGB}{190,80,120}
\definecolor{lightyellow}{RGB}{255,253,208}
\definecolor{darkyellow}{RGB}{204,163,82}
\definecolor{lightorange}{RGB}{255,240,220}
\definecolor{darkorange}{RGB}{230,120,30}
\definecolor{lightgreen}{RGB}{220,245,230}
\definecolor{darkgreen}{RGB}{90,150,100}

\tikzstyle{my-box}=[
    rectangle,
    draw=hidden-draw,
    rounded corners,
    text opacity=1,
    minimum height=1.5em,
    minimum width=5em,
    inner sep=2pt,
    align=center,
    fill opacity=.5,
    line width=0.8pt,
]
\tikzstyle{leaf}=[my-box, minimum height=1.5em,
    fill=hidden-pink!80, text=black, align=left,font=\normalsize,
    inner xsep=5pt,
    inner ysep=4pt,
    line width=0.8pt,
]
\begin{figure*}[t!]
    \centering
    \resizebox{\textwidth}{!}{
        \begin{forest}
            forked edges,
            for tree={
                grow=east,
                reversed=true,
                anchor=base west,
                parent anchor=east,
                child anchor=west,
                base=center,
                font=\large,
                rectangle,
                draw=hidden-draw,
                rounded corners,
                align=left,
                text centered,
                minimum width=4em,
                edge+={darkgray, line width=1pt},
                s sep=3pt,
                inner xsep=2pt,
                inner ysep=3pt,
                line width=0.8pt,
                ver/.style={rotate=90, child anchor=north, parent anchor=south, anchor=center},
            },
            where level=1{text width=10em,font=\normalsize,}{},
            where level=2{text width=12em,font=\normalsize}{},
            where level=3{text width=8em,font=\normalsize,}{},
            where level=4{text width=6em,font=\normalsize,}{},
            [
                {Latent Chain-of-Thought Reasoning}, ver, text width=18em, align=center, fill=maingray, draw=darkgray, line width=1pt
                [
                    {Token-wise \\ Horizontal Level (\S\ref{sec:Token-wise Strategies})}, align=center, fill=lightpurple, draw=darkpurple, line width=0.8pt
                    [
                        {Representation \\ Initialization (\S\ref{sec:representationini})}, align=center, fill=lightpurple, draw=darkpurple, line width=0.8pt
                            [
                                {\underline{\textit{Hidden State}}: e.g.,}
                                {Coconut~\citep{hao2024traininglargelanguagemodels}}{, }
                                {CODI~\citep{shen2025codi}}{, }
                                {LatentSeek~\citep{li2025seekdarkreasoningtesttime}}{, }\\
                                {System-1.5 Reasoning~\citep{wang2025system15reasoningtraversallanguage}}{, }
                                {PCCoT~\citep{wu2025parallelcontinuouschainofthoughtjacobi}}{, }
                                {KaVa~\citep{kuzina2025kavalatentreasoningcompressed}}{, }\\
                                {FR-Ponder~\citep{he2025learningponderadaptivereasoning}}{, }
                                {R-Capsule}{~\citep{shan2025rcapsulecompressinghighlevelplans}}{, }
                                {SIM-CoT~\citep{wei2025simcotsupervisedimplicitchainofthought}}{, }
                                {Pythia Arch}\\{~\citep{zeng2025pretrainingllmlatentthoughts}}{, }
                                {CCoT}{~\citep{cheng2024compressed}}{, }
                                {HCoT~\citep{liu2024expeditingelevatinglargelanguage}}{, }
                                {LatentEvolve}\\{~\citep{zhang2025latentevolveselfevolvingtesttimescaling}}{, }
                                {LTA-Thinker~\citep{wang2025ltathinkerlatentthoughtaugmentedtraining}}{, }
                                {SoftCoT~\citep{xu2025softcotsoftchainofthoughtefficient,xu2025softcottesttimescalingsoft}}, 
                                % {SoftCoT++}{~\citep{xu2025softcottesttimescalingsoft}}, 
                                % {~\citet{wang2025efficientposttrainingrefinementlatent}},
                                leaf, text width=44.2em, fill=lightpurple, draw=none
                            ]
                            [
                                {\underline{\textit{Weighted Embedding}}: e.g.,}
                                {Soft Thinking~\citep{zhang2025softthinkingunlockingreasoning}}{, }
                                {HRPO~\citep{yue2025hybridlatentreasoningreinforcement}}{, }
                                {MoT-G}\\{~\citep{jain2025learningreasonmixturetokens}}{, }
                                {CoLaR~\citep{tan2025thinksilentlythinkfast}}{, }
                                {\citet{butt2025softtokenshardtruths}},
                                leaf, text width=44.2em, fill=lightpurple, draw=none
                            ]
                            [
                                {\underline{\textit{Special Vector}}: e.g.,}
                                {Token Assorted~\citep{su2025token}}{, }
                                {GainRouter~\citep{zheng2025fastthinkinglargelanguage}}{, }
                                {Latent Tokens}\\{~\citep{sun2025enhancinglatentcomputationtransformers}}{, }
                                {LightThinker~\citep{zhang2025lightthinkerthinkingstepbystepcompression}}{, }
                                {CoCoMix~\citep{tack2025cocomix}}{, }
                                {DART}\\{~\citep{jiang2025dartdistillingautoregressivereasoning}}{, }
                                {SynAdapt}{~\citep{wang2025synadaptlearningadaptivereasoning}}{, }
                                {MarCos~\citep{liu2025marcosdeepthinkingmarkov}},
                                leaf, text width=44.2em, fill=lightpurple, draw=none
                            ]
                    ]
                    [
                        {Model Optimization (\S\ref{sec:modelopt})}, fill=lightpurple, draw=darkpurple, line width=0.8pt
                        [
                            {\underline{\textit{Pre-Training}}: e.g.,}
                            {CoCoMix~\citep{tack2025cocomix}}{, }
                            {Ponder-LM2~\citep{zeng2026ponderlm}},
                            leaf, text width=44.2em, fill=lightpurple, draw=none
                        ]
                        [
                            {\underline{\textit{Post-Training}}:}
                            {\textit{Supervised fine-tuning:} e.g.,}
                            {Coconut~\citep{hao2024traininglargelanguagemodels}}{, }
                            {CODI~\citep{shen2025codi}}{, }\\
                            {PCCoT}{~\citep{wu2025parallelcontinuouschainofthoughtjacobi}}{, }
                            {HCoT~\citep{liu2024expeditingelevatinglargelanguage}}{, }
                            {CCoT~\citep{cheng2024compressed}}{, }
                            {LTA-Thinker}\\{~\citep{wang2025ltathinkerlatentthoughtaugmentedtraining}}{, }
                            {CoLaR~\citep{tan2025thinksilentlythinkfast}}{, }
                            {System-1.5 Reasoning~\citep{wang2025system15reasoningtraversallanguage}}{, }
                            {SynAdapt}\\{~\citep{wang2025synadaptlearningadaptivereasoning}}{, }
                            {KaVa~\citep{kuzina2025kavalatentreasoningcompressed}}{, }
                            {MarCos~\citep{liu2025marcosdeepthinkingmarkov}}{, }
                            {R-Capsule}\\{~\citep{shan2025rcapsulecompressinghighlevelplans}}{, }
                            {SIM-CoT~\citep{wei2025simcotsupervisedimplicitchainofthought}}
                            {;}
                            {\textit{Reinforcement Learning:} e.g.,}
                            {CoLaR~\citep{tan2025thinksilentlythinkfast}}{, }\\
                            {HRPO~\citep{yue2025hybridlatentreasoningreinforcement}}{, }
                            {LatentSeek~\citep{li2025seekdarkreasoningtesttime}}{, }
                            {FR-Ponder~\citep{he2025learningponderadaptivereasoning}}{, }\\
                            {MoT-G~\citep{jain2025learningreasonmixturetokens}}{, }
                            {\citet{butt2025softtokenshardtruths}}{, }
                            {SLPO~\citep{you2026slposcalinglatentreasoning}}, 
                            leaf, text width=44.2em, fill=lightpurple, draw=none
                        ]
                    ]
                    [
                        {Inference Exploration (\S\ref{sec:inferenceexp})}, fill=lightpurple, draw=darkpurple, line width=0.8pt
                        [
                            {\underline{\textit{Sequential Scaling}}: e.g.,}
                            {LatentSeek~\citep{li2025seekdarkreasoningtesttime}}{, }
                            {System-1.5 Reasoning~\citep{wang2025system15reasoningtraversallanguage}}{, }\\
                            {LatentEvolve~\citep{zhang2025latentevolveselfevolvingtesttimescaling}}{, }
                            {FR-Ponder~\citep{he2025learningponderadaptivereasoning}},
                            leaf, text width=44.2em, fill=lightpurple, draw=none
                        ]
                        [
                            {\underline{\textit{Parallel Scaling}}: e.g.,}
                            {SoftCoT++~\citep{xu2025softcottesttimescalingsoft}}{, }
                            {PCCoT~\citep{wu2025parallelcontinuouschainofthoughtjacobi}}{, }
                            {\citet{butt2025softtokenshardtruths}}{, }\\
                            {KaVa~\citep{kuzina2025kavalatentreasoningcompressed}}{, }
                            {LTA-Thinker~\citep{wang2025ltathinkerlatentthoughtaugmentedtraining}}{, }
                            {Pythia Arch~\citep{zeng2025pretrainingllmlatentthoughts}}, 
                            leaf, text width=44.2em, fill=lightpurple, draw=none
                        ]
                    ]
                ]
                [
                    {Layer-wise \\ Vertical Level (\S\ref{sec:Internal Mechanism})}, align=center, fill=lightblue, draw=darkblue, line width=0.8pt
                    [
                        {Model Architectures}, fill=lightblue, draw=darkblue, line width=0.8pt
                        [
                            {\underline{\textit{Encoder-Based Models}}: e.g.,}
                            {RELAY~\citep{RELAY}}{, }
                            {HRM~\citep{Hierarchical_Reasoning_Model}},
                            leaf, text width=44.2em, fill=lightblue, draw=none
                        ]
                        [
                            {\underline{\textit{Decoder-Based Models}}: e.g.,}
                            {CoTFormer~\citep{mohtashami2024cotformerchainofthoughtdrivenarchitecture}}{, }
                            {Huginn~\citep{scaling_up_test_time}}{, }
                            {LTO}\\{~\citep{du2025latentthinkingoptimizationlatent}}{, }
                            {ITT~\citep{inner_thinking_transformer}}{, }
                            {Pondering LM~\citep{zeng2025pretraininglanguagemodelsponder}}{, }\\
                            {\citet{zhu20254thdimensionscalingmodel}}{, }
                            {MoR~\citep{bae2025mixtureofrecursionslearningdynamicrecursive}},
                            leaf, text width=44.2em, fill=lightblue, draw=none
                        ]
                    ]
                ]
                [
                    {Analysis (\S\ref{sec:analysis})}, align=center, fill=lightorange, draw=darkorange, line width=0.8pt
                        [
                            {Behavioral Observations(\S\ref{sec:ext-behavior})}, fill=lightorange, draw=darkorange, line width=0.8pt
                            [
                            {e.g.,}
                                {\citet{li2025skiplayerloopit}}{, }
                                {\citet{liu2024languagemodelslearnskip}}{, }
                                {\citet{lin2025implicit}}{, }
                                {\citet{zhu20254thdimensionscalingmodel}}{, }
                                {\citet{saunshi2025looped}},
                                leaf, text width=44.2em, fill=lightorange, draw=none
                            ]
                        ]
                        [
                            {Internal Mechanisms (\S\ref{sec:Internal_Mechanisms})}, align=center, fill=lightorange, draw=darkorange, line width=0.8pt
                            [
                            {\underline{\textit{Empirical Analysis}}: e.g.,}
                                {\citet{lindsey2025biology}}{, }
                                {\citet{hou-etal-2023-towards}}{, }
                                \citet{lin2025implicit}{, }
                                \citet{yang2024largelanguagemodelslatently}{, }\\
                                \citet{shalev2024distributionalreasoningllmsparallel}{, }
                                \citet{wang2024grokkedtransformersimplicitreasoners}{, }
                                \citet{kudo2025thinktotalktalktothinkllmscome}{, }
                                \citet{yu2025llmsreallythinkstepbystep}{, }
                                \citet{zhang2025uncoveringlatentchainthought}{, }\\
                                \citet{brinkmann2024mechanisticanalysistransformertrained}{, }
                                \citet{wang2025latentspacechainofembeddingenables}{, }
                                \citet{yang2025internalcot}{, }
                                \citet{lin2025identitybridgeenablingimplicit}{, }
                                \citet{wu2025llmssinglethreadedreasonersdemystifying}{, }\\
                                \citet{csordás2025languagemodelsusedepth}{, }
                                \citet{li2024understandingpatchingcompositionalreasoning},
                                leaf, text width=44.2em, fill=lightorange, draw=none
                            ]
                            [
                            {\underline{\textit{Theoretical Analysis}}: e.g.,}
                                \citet{zhu2025reasoningsuperpositiontheoreticalperspective}{, }
                                \citet{xu2025formalcomparisonchainofthoughtlatent}{, }
                                \citet{gozeten2025continuouscot}{, }\\
                                \citet{zhu2025emergencesuperpositionunveilingtraining}{, }
                                \citet{saunshi2025looped}{, }\citet{chen2026what},
                                leaf, text width=44.2em, fill=lightorange, draw=none
                            ]
                        ]                       
                ]
                [
                    {Applications (\S\ref{sec:application})}, fill=lightgreen, draw=darkgreen, line width=0.8pt
                    [   
                    {e.g.,}
                        {Heima~\citep{shen2025efficientreasoninghiddenthinking}}{, }
                        {LatentLM~\citep{sun2024multimodall}}{, }{, }
                        {SSR~\citep{liu2025ssrenhancingdepthperception}}{, }
                        {Mirage~\citep{yang2025machinementalimageryempower}}{, }
                        {MCOUT}\\{~\citep{pham2025multimodalchaincontinuousthought}}{, }
                        {LVR~\citep{li2025latentvisualreasoning}}{, }
                        {Debater~\citep{ji2025learningeffectiverepresentationsdense}}{, }
                        {ReaRec~\cite{tang2025thinkrecommendunleashinglatent}}{, }
                        {LCR-SER~\citep{shi2025bridgingsearchrecommendationlatent}}{, }\\
                        {LARES~\citep{liu2025lareslatentreasoningsequential}}{, }
                        {LatentR3~\citep{zhang2025reinforcedlatentreasoningllmbased}}{, }
                        {LCDrive~\citep{tan2025latentchainofthoughtworldmodeling}}{, }
                        {LatentMAS~\citep{zou2025latentcollaborationmultiagentsystems}},
                        leaf, text width=58em, fill=lightgreen, draw=none
                    ]
                ]
            ]
        \end{forest}
     }
    % \vspace{-16pt}
    \caption{Taxonomy of latent Chain-of-Thought (CoT) reasoning.}
    \label{fig:taxo_of_LCoT}
    \vspace{-12pt}
\end{figure*}

However, the intrinsic constraints in natural language format limits LLM reasoning capabilities. The first is \textbf{expressive redundancy}. Many tokens in a reasoning chain are syntactically necessary but functionally non-essential to the reasoning process (e.g., ``so,'' ``the''), which inflates token usage and slows inference without proportionate gains in reasoning quality~\citep{feng2025efficientreasoningmodelssurvey, sui2025stopoverthinkingsurveyefficient, shen2025codi}. This redundancy also increases the chance of overfitting to stylistic artifacts rather than genuine reasoning signals~\cite{lippmann2025style}. The second is a \textit{semantic bottleneck}. Human cognition often transcends discrete linguistic symbols, involving abstract, continuous, or multi-conceptual representations that resist precise verbalization~\cite{wittgenstein1922tractatus}.\footnote{For instance, complex emotions like nostalgia and bittersweetness are continuous blends of joy, sadness, and memory that defy expressions from a fixed vocabulary~\citep{ pinker1994language}.} Forcing continuous reasoning dynamics into a linear chain of fixed vocabulary inevitably leads to information loss~\cite{hao2024traininglargelanguagemodels}. 

These intrinsic constraints have driven a growing shift toward \textbf{latent Chain-of-Thought (latent CoT)} reasoning, where models reason not through explicit language tokens but within continuous latent spaces~\citep{zhang2025latentevolveselfevolvingtesttimescaling}. As illustrated in Figure~\ref{fig:intro}, latent CoT replaces linguistic articulation with dense, high-dimensional representations that serve as a more native substrate for computational reasoning. This ``de-linguistified'' paradigm offers several advantages: it accelerates inference by reducing token-level computation~\citep{liu2025marcosdeepthinkingmarkov}, allows for richer and more compact reasoning representations~\citep{li2025seekdarkreasoningtesttime, butt2025softtokenshardtruths}, and enables the parallel exploration of multiple reasoning trajectories~\citep{butt2025softtokenshardtruths}.

Nonetheless, this emerging paradigm presents new challenges, primarily stemming from the unobservable nature of the reasoning process. This creates a significant \textit{training and alignment} problem, making it difficult to apply direct supervision~\citep{xu2025softcotsoftchainofthoughtefficient, ruan2025reasoning, lindsey2025biology}. Without proper supervision, latent trajectories may drift, fail to develop structured internal reasoning or ensure ethical controllability. Furthermore, the lack of transparency leads to an \textit{evaluation gap}: it is unclear whether models are performing genuine reasoning or simply exploiting input–output correlations~\citep{ameisen2025circuit}.

To address these open questions, research on latent CoT has seen rapid and enthusiastic growth. Recent efforts can be broadly categorized into several directions: some works focus on novel architectural designs~\citep{zhu2025surveylatentreasoning} and implicit reasoning~\citep{li2025implicitreasoninglargelanguage}, while others explore latent CoT as a vehicle for efficient or compressed reasoning~\citep{sui2025stopoverthinkingsurveyefficient, feng2025efficientreasoningmodelssurvey}. Given the field’s accelerating pace and expanding scope, there is a growing need for a comprehensive and structured overview. This work provides the first in-depth survey of latent CoT, offering a unified taxonomy that organizes the current landscape in terms of methodological foundations, training paradigms,  and cognitive implications.

Our key contributions are threefold:
(1) \textbf{Systematic Taxonomy.} We propose the first unified taxonomy that organizes latent CoT studies into two primary methodological categories (Figure~\ref{fig:taxo_of_LCoT}), clarifying their assumptions, objectives, and innovations.
(2) \textbf{In-depth Discussion.} Building on this taxonomy, we conduct a detailed analysis of representative works, comparing training strategies, architectural paradigms, supervision signals, and efficiency trade-offs to provide a consolidated view of the design space.
(3) \textbf{Challenge and Future Directions.} We identify critical open problems in latent CoT—such as unsupervised training, evaluation faithfulness, and interpretability—and outline promising directions for future research.
\section{Preliminaries}

\paragraph{Definitions}
We begin by formalizing the key concepts central to this survey. 
A \textbf{Chain-of-Thought (CoT)} is a sequence of intermediate reasoning steps, $c = (c_1, c_2, \dots, c_T)$, that connects an input query $x$ to its final answer $y$~\citep{wei2022chain}. This reasoning process can be instantiated in two primary forms, distinguished by the medium in which the “thoughts” unfold.
The first is \textbf{explicit CoT}, where the reasoning is verbalized in natural language. Each step $c_t$ is materialized as a sequence of discrete tokens from a vocabulary $\mathcal{V}$ (i.e., $c_t \in \mathcal{V}^*$). This approach renders the model's thought process transparent and human-readable.
In contrast, \textbf{Latent CoT} represents the reasoning trace in a non-verbal, high-dimensional latent space~\citep{scaling_up_test_time}. Here, each step $c_t$ corresponds to an compact computational structure, such as a dense vector or a transformation of the model's hidden states. This approach offers a more compact and potentially more expressive medium for complex cognitive processes, moving beyond the constraints of discrete language~\citep{hao2024traininglargelanguagemodels}.

\paragraph{Overview}
This survey systematically organizes the field of latent CoT reasoning. Our taxonomy (Figure~\ref{fig:taxo_of_LCoT}) is structured along two primary axes, inspired by the distinction between horizontal and vertical computation in Transformers~\citep{deng2023distillcot}. The first axis, the \textbf{token-wise horizontal level}~(\S\ref{sec:Token-wise Strategies}), covers methods that generate a sequence of latent thoughts before producing the final answer. 
% Our review of this area is structured around the key stages of representation initialization, optimization, and inference exploration. 
The second axis, the \textbf{layer-wise vertical level}~(\S\ref{sec:Internal Mechanism}), focuses on methods that deepen reasoning through iterative computation across layers, which we categorize by model architecture. Beyond this methodological core, the survey covers key \textbf{analyses}~(\S\ref{sec:analysis}), real-world \textbf{applications}~(\S\ref{sec:application}), and concludes with \textbf{challenges and future directions}~(\S\ref{sec:challenges}). 

\section{Token-wise Horizontal Level}
\label{sec:Token-wise Strategies}

\begin{table*}[htbp]
\centering
\resizebox{\textwidth}{!}{
\begin{tabular}{llcl}
\toprule
\textbf{Backbone} & \textbf{Method} & \textbf{GSM8K Acc. (\%)} & \textbf{\#Tokens (avg) / efficiency note} \\
\midrule
\multirow{8}{*}{\textbf{GPT-2}} 
 & No-CoT-SFT* & 19.1 & 2.2 \\
 & CoT-SFT* & 41.8 & 25 \\
 & Coconut~\citep{hao2024traininglargelanguagemodels} & 34.1 & 8.2 (6 continuous thought tokens) \\
 & CODI~\citep{shen2025codi} & 43.7 & N/R (6 continuous thought tokens + answer tokens) \\
 & PCCoT~\citep{wu2025parallelcontinuouschainofthoughtjacobi} & 49.48 & N/R (24 continuous thought, $T=3$ extra iterations) \\
 & SIM-CoT~\citep{wei2025simcotsupervisedimplicitchainofthought} (+Coconut) & 44.8 & 12.2 \\
 & SIM-CoT~\citep{wei2025simcotsupervisedimplicitchainofthought} (+CODI) & 42.6 & 12.2 \\
 & System-1.5 Reasoning~\citep{wang2025system15reasoningtraversallanguage} & 46.66 & N/R (paper reported \#Steps=2; CoT-SFT \#Steps=26) \\
\midrule
\multirow{9}{*}{\textbf{LLaMA 3-1B-it}} 
 & No-CoT-SFT* & 44.1 & 2.2 \\
 & CoT-SFT* & 66.7 & 25 \\
 & Coconut*~\citep{hao2024traininglargelanguagemodels} & 45.3 & 8.2 (6 continuous thought tokens) \\
 & CODI~\citep{shen2025codi} & 55.6 & N/R (6 continuous thought tokens + answer tokens) \\
 & PCCoT~\citep{wu2025parallelcontinuouschainofthoughtjacobi} & 53.35 & N/R (24 continuous thought, $T=3$ extra iterations) \\
 & SIM-CoT~\citep{wei2025simcotsupervisedimplicitchainofthought} (+Coconut) & 42.2 & 13.2 \\
 & SIM-CoT~\citep{wei2025simcotsupervisedimplicitchainofthought} (+CODI) & 56.1 & 13.2 \\
 & KaVa~\citep{kuzina2025kavalatentreasoningcompressed} & 55.07 & N/R (6.9 forward passes required for trace and answer) \\
 & CoLaR~\citep{tan2025thinksilentlythinkfast} & 40.1 & 12.7 \\
\bottomrule
\end{tabular}}
\caption{Comparison of performance and efficiency across representative methods. Results are sourced from their respective original papers. ``N/R'' indicates metrics not reported in the original papers. * denotes baseline results reproduced and reported by CODI.}
\label{tab:performance_efficiency}
\end{table*}

Methods at the \textbf{token-wise horizontal level} are characterized by the generation of intermediate latent thoughts along the sequence dimension to guide the reasoning process. This approach allows the model to ``think ahead'' by forming a coherent reasoning plan in a compact, latent format, rather than verbalizing every step.

The intellectual roots of this paradigm can be traced to two complementary lines of pioneering research. One line explored giving models an explicit budget for extra computation by inserting designated “thinking” or “planning” tokens into the sequence~\citep{goyal2023think, wang2023guiding, herel2024thinking}. A parallel thread aimed to internalize this reasoning process, embedding the logic of an explicit CoT directly into the model’s latent representations without extending the visible token sequence~\citep{deng2023distillcot, deng2024explicit, xu2025twtthinkingtokenshabitual}.\footnote{Appendix~\ref{sec:pioneerwork} provides details on this pioneering work.}
Both of these early threads converged on a key insight: the primary benefit stems from enabling a rich \textit{computational process} within the model, independent of the explicit \textit{linguistic content} generated at inference time. 
Following a common conceptualization of Transformer computation which distinguishes \texttt{horizontal} processing along the sequence from \texttt{vertical} processing across layers~\citep{deng2023distillcot}, we define methods that generate intermediate latent thoughts to guide the reasoning process before producing the tokens of the final answer as \textit{token-wise horizontal level}, and we structure our discussion around three key stages: \textbf{representation initialization}, \textbf{model optimization}, and \textbf{inference exploration}.

\subsection{Representation Initialization}
\label{sec:representationini}

\begin{figure}[t]
\centering
\includegraphics[width=1.0\columnwidth]{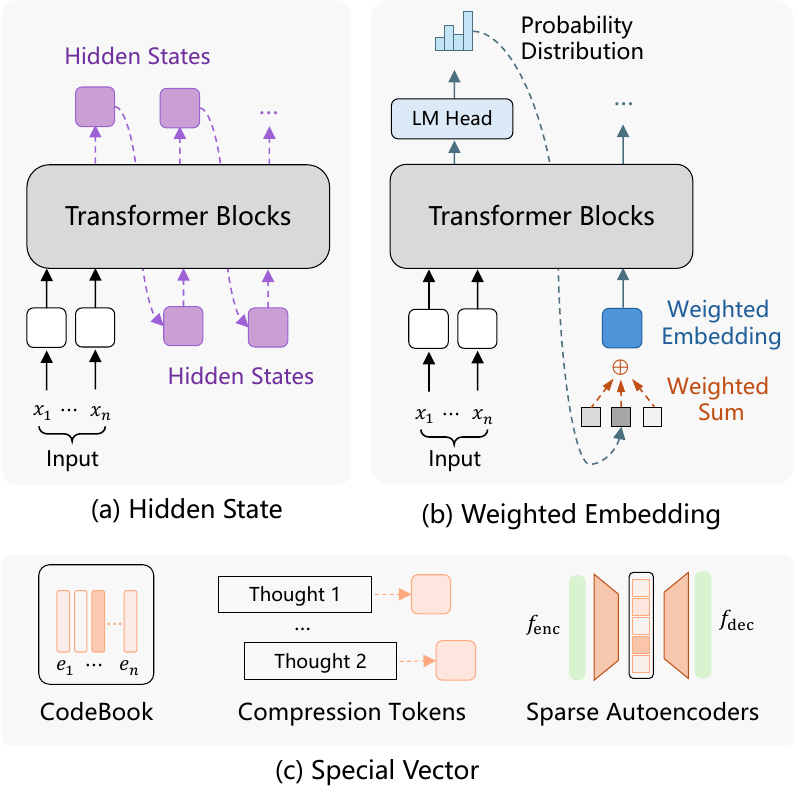}
    \caption{Illustration of three representative categories for latent representation initialization. 
    }
    \vspace{-12pt}
    \label{fig:representation}
\end{figure}

A crucial first step is determining how the latent thoughts are represented. This \textit{Representation Initialization} is a foundational design choice, as it dictates the nature of the latent space and strongly influences the model's capacity to reason. As shown in Figure~\ref{fig:representation}, we survey existing works by categorizing it into three main types based on the origin of the latent representation: model's internal \emph{hidden states}, \emph{weighted embedding} of existing vocabulary tokens, and newly introduced \emph{special vectors}\footnote{Appendix~\ref{sec:special_vector} provides details on these methods. A detailed comparison of representative methods across these categories, including their optimization objectives, model backbones, and training and evaluation datasets, is provided in Appendix Table~\ref{tab:technical_comparison}.}. 

\paragraph{Hidden State}
One of the most direct and efficient strategies for initialization is to derive latent thoughts from the model's hidden states. This approach leverages the rich, contextualized representations that already exist within the model, treating them as a native medium for thought. Current methods can be broadly divided based on whether the hidden states originate from the model itself or an auxiliary model. Within the \textit{self-contained} approach, a prevalent technique is to establish a sequential state-passing mechanism. Works like \textit{Chain of Continuous Thought}, ~\citep{hao2024traininglargelanguagemodels}, \textit{Pythia Arch}~\citep{zeng2025pretrainingllmlatentthoughts}, \textit{KV-cache distillation}~\citep{kuzina2025kavalatentreasoningcompressed} and \textit{System-1.5 Reasoning}~\citep{wang2025system15reasoningtraversallanguage} take the final hidden state from a given step and re-inject it as the input for the subsequent step, effectively replacing the standard token embedding~\citep{wang2025efficientposttrainingrefinementlatent}. 
To better align the output hidden state with the model's input space, some approaches like \textit{Continuous Chain-of-Thought via Self-Distillation}~\citep{shen2025codi}, \textit{Parallel Continuous Chain-of-Thought}~\citep{wu2025parallelcontinuouschainofthoughtjacobi} and \textit{Supervised IMplicit-CoT}~\citep{wei2025simcotsupervisedimplicitchainofthought} pass the previous hidden state through a projection network before the next step. 
Other variations involve using a special token as a trigger~\citep{shan2025rcapsulecompressinghighlevelplans}, extracting and refining a full sequence of hidden states~\citep{li2025seekdarkreasoningtesttime}, or applying additive updates to the hidden state~\citep{he2025learningponderadaptivereasoning}.
Alternatively, the \textit{auxiliary model} approach delegates the generation of the latent thought to a separate model which then provides a guiding hidden state back to the main LLM~\citep{liu2024expeditingelevatinglargelanguage, cheng2024compressed, xu2025softcotsoftchainofthoughtefficient, xu2025softcottesttimescalingsoft, zhang2025latentevolveselfevolvingtesttimescaling, wang2025ltathinkerlatentthoughtaugmentedtraining}.
%These methods treat the model's hidden states as a dynamic internal scratchpad, with the primary design choice revolving around how information is propagated through the latent tokens or steps.

\paragraph{Weighted Embedding}
This strategy operates directly in the \textit{model's input embedding space} rather than its output hidden state space. By keeping the computation within the embedding space, it ensures that latent thoughts remain highly tied to the model's learned vocabulary~\citep{zhang2025softthinkingunlockingreasoning, zhuang2025textgen, yue2025hybridlatentreasoningreinforcement, butt2025softtokenshardtruths, jain2025learningreasonmixturetokens, shi2025swireasoningswitchthinkinglatentexplicit, deng2025latentreasoningllmsvocabularyspace}. It creates a superposition of concepts calculated as a weighted average of existing vocabularies, which allows the model to capture uncertainty and nuance beyond a single output.

\subsection{Model Optimization}
\label{sec:modelopt}

With latent representation initialized, we shift the focus to model optimization. 
Since methods vary primarily in computational phases, we organize them into two main paradigms: (1) \emph{Pre-Training} methods that integrate latent reasoning into the model's pre-training stage, and (2) \emph{Post-Training} techniques, which fine-tune base models via specialized datasets and objectives. For reference, we summarize the performance and computational efficiency of representative methods in Table~\ref{tab:performance_efficiency}.

\subsubsection{Pre-Training}
The central idea is to modify the standard language modeling objective so that the model must perform an intermediate reasoning computation, within latent space, before producing each token prediction. Recent studies demonstrate that this integration can be achieved along a spectrum of supervision. \textit{PonderLM-2}~\citep{zeng2026ponderlm} claims that models can learn latent reasoning implicitly, where the intermediate thoughts emerge naturally as a byproduct of optimizing next-token prediction. \textit{Continuous Concept Mixing}~\citep{tack2025cocomix} explicitly guided reasoning process by incorporating auxiliary semantic signal. 
However, such approaches also introduce substantial computational overhead and make it difficult to disentangle whether observed improvements stem from genuine reasoning ability or from additional representational capacity.

\subsubsection{Post-Training} 
Within the dominant Post-Training paradigm, we further distinguish between methods based on \emph{Supervised Fine-Tuning (SFT)} and those leveraging \emph{Reinforcement Learning (RL)}. SFT-based approaches train the model to produce latent thoughts by providing it with supervision signals. These methods can be categorized by whether the supervision on the latent process is \textit{indirect}, focusing on the final outcome, or \textit{direct}, targeting the latent representations themselves.
\paragraph{Indirect Supervision in SFT.} This class of methods optimizes latent reasoning by applying a loss function only to the human-readable output, without providing an direct target for the intermediate latent thoughts, claiming that by learning to predict the correct subsequent text, the model will \textit{implicitly form useful latent representations}~\citep{hao2024traininglargelanguagemodels}. Another form of indirect guidance is provided through knowledge distillation to constrain certain token's hidden representation and thus align the student model's behavior with that of a teacher model that performs explicit CoT~\citep{shen2025codi, wu2025parallelcontinuouschainofthoughtjacobi}. ~\citet{wang2025ltathinkerlatentthoughtaugmentedtraining} introduce distributional constraints, using KL divergence to ensure representation remain semantically close to the question, contrastive learning to focus them on critical reasoning steps, or a VAE-like framework to regularize representation by matching a prior distribution to a posterior inferred from the ground-truth text~\citep{liu2025marcosdeepthinkingmarkov}.
\paragraph{Direct Supervision in SFT.}
These methods provide an \textit{explicit ``gold'' target for the latent representations}, typically derived from the content they are intended to replace. This offers greater control over the latent process and can improve training stability. One popular strategy is to train the representation to \textit{reconstruct their corresponding linguistic reasoning steps}. This is often implemented via an auxiliary decoder that takes each latent vector as input and is trained with a standard cross-entropy loss to reproduce the original text of that reasoning step or a higher-level textual plan~\citep{wei2025simcotsupervisedimplicitchainofthought, he2025semcotacceleratingchainofthoughtreasoning, shan2025rcapsulecompressinghighlevelplans, he2025semcotacceleratingchainofthoughtreasoning}. Other methods use a contrastive loss to pull the representation of a special latent token closer to a pooled representation of the full text it represents~\citep{liu2024expeditingelevatinglargelanguage}. More direct approaches use a Mean Squared Error loss to train a student model to generate latent vectors that numerically match the teacher's final hidden states at corresponding steps~\citep{cheng2024compressed, wang2025system15reasoningtraversallanguage}. The most granular form of this supervision involves providing a dense, step-wise signal by training the student's entire latent KV-cache to mimic a compressed version of the teacher's KV-cache~\citep{kuzina2025kavalatentreasoningcompressed} or supervising a latent head to predict the next compressed embedding in the reasoning chain~\citep{tan2025thinksilentlythinkfast}.
\paragraph{Reinforcement Learning.}
RL provides an alternative to SFT that enables models to autonomously discover reasoning strategies. Instead of being guided by ground-truth CoTs, the model is trained to \textit{maximize a reward signal} that is typically based on the final outcome. A common approach is to first initialize the model with SFT and then fine-tune it using policy gradient algorithms, such as REINFORCE or the more sample-efficient Group Relative Policy Optimization (GRPO)~\citep{tan2025thinksilentlythinkfast, butt2025softtokenshardtruths, jain2025learningreasonmixturetokens, butt2025softtokenshardtruths}. To ensure training stability, a KL divergence penalty is often used to keep the model from deviating excessively from a reference policy~\citep{yue2025hybridlatentreasoningreinforcement, butt2025softtokenshardtruths}. This framework can be extended beyond simple accuracy rewards to incorporate other objectives like computational efficiency~\citep{he2025learningponderadaptivereasoning}.

\begin{figure}[t]
\centering
\includegraphics[width=0.8\columnwidth]{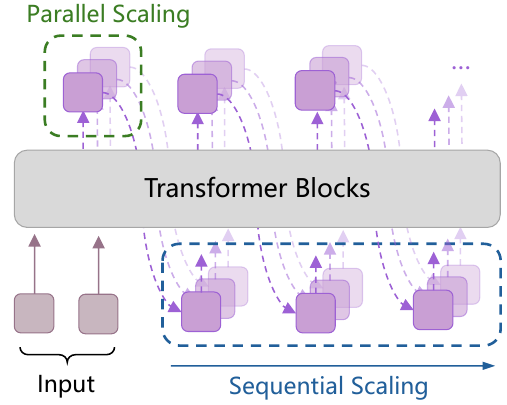}
    \caption{Illustration of inference exploration (both parallel and sequential scaling) for latent CoT reasoning. 
    }
    \vspace{-12pt}
    \label{fig:exploration}
\end{figure}

\subsection{Inference Exploration}
\label{sec:inferenceexp}
Beyond model training, the full potential of latent reasoning models can be further unlocked during inference~\citep{wang2025inferencetimescalingcontinuousspace}. Following test-time scaling paradigms~\citep{wang2025scalingscalingexploringtesttime, zhang2025surveytesttimescalinglarge}, we divide inference exploration strategies into two types (as illustrated in Figure~\ref{fig:exploration}): 
(1) \emph{Sequential Scaling} methods, which generate latent thoughts one after another to form a single reasoning chain. They focus on refining a single chain of latent thoughts. These techniques can improve reasoning quality by iteratively optimizing the latent trajectory at test-time~\citep{li2025seekdarkreasoningtesttime, zhang2025latentevolveselfevolvingtesttimescaling, ye2025thinkingflytesttimereasoning} or enhance efficiency by adaptively allocating computational resources based on problem difficulty~\citep{wang2025system15reasoningtraversallanguage, he2025learningponderadaptivereasoning}. (2) \emph{Parallel Scaling} methods, which explore multiple reasoning paths simultaneously to improve solution quality or efficiency~\citep{you2025paralleltesttimescalinglatent}. This approach boosts performance and robustness by exploring a diverse set of potential solutions, analogous to self-consistency in explicit reasoning~\citep{xu2025softcottesttimescalingsoft, butt2025softtokenshardtruths, wu2025parallelcontinuouschainofthoughtjacobi}. These strategies can also be designed to improve inference latency through concurrent processing of reasoning steps~\citep{wang2025ltathinkerlatentthoughtaugmentedtraining, zeng2025pretrainingllmlatentthoughts}.

\section{Layer-wise Vertical Level}
\label{sec:Internal Mechanism}

Inspired by findings that human reasoning often involves recurrent neural activity before verbal expression~\citep{GOLDMANRAKIC1995477}, recent studies augment LLMs with looping units that \textit{iteratively refine hidden states through layer-wise vertical loops}, allowing the model’s internal state to consolidate a ``thought'' before producing the next token, as illustrated in Figure~\ref{fig:structural_cot}.
This \textit{vertical, loop-driven} mechanism is designed to emulate a human-like “think-then-speak” dynamic—preserving the latent-space style of reasoning while still maintaining concise internal traces of intermediate progress—thereby offering a middle ground between fully verbalized CoT and purely silent token-wise reasoning. 
We categorize existing methods according to the type of architecture: \emph{encoder-based} versus \emph{decoder-based} Transformers.

\paragraph{Encoder-based Models}  
This category refers to approaches that adopt an \textit{encoder-only looped Transformer} as the core reasoning engine, where the encoder is repeatedly applied to iteratively refine the latent representation through vertical loop iterations.  
The \textit{Hierarchical Reasoning Model}~\citep{Hierarchical_Reasoning_Model}, for instance, employs a two-tier structure to enable coarse-to-fine reasoning across loops without generating explicit thought steps. 
\textit{REasoning through Loop Alignment iterativelY}~\citep{RELAY} builds on an encoder-based looped Transformer with strong length generalization by aligning each loop iteration to a CoT step, enabling it to produce reliable long-horizon reasoning chains that are subsequently used to train auto-regressive models.

\paragraph{Decoder-based Models} In contrast to encoder-looped reasoning, these approaches employ a \textit{loop-driven decoder} that repeatedly re-applies (part of) the decoder stack to refine the hidden state before emitting each output token. Existing methods can be broadly divided into two categories: 
\begin{figure}[t]
    \centering
    \includegraphics[width=0.98\linewidth]{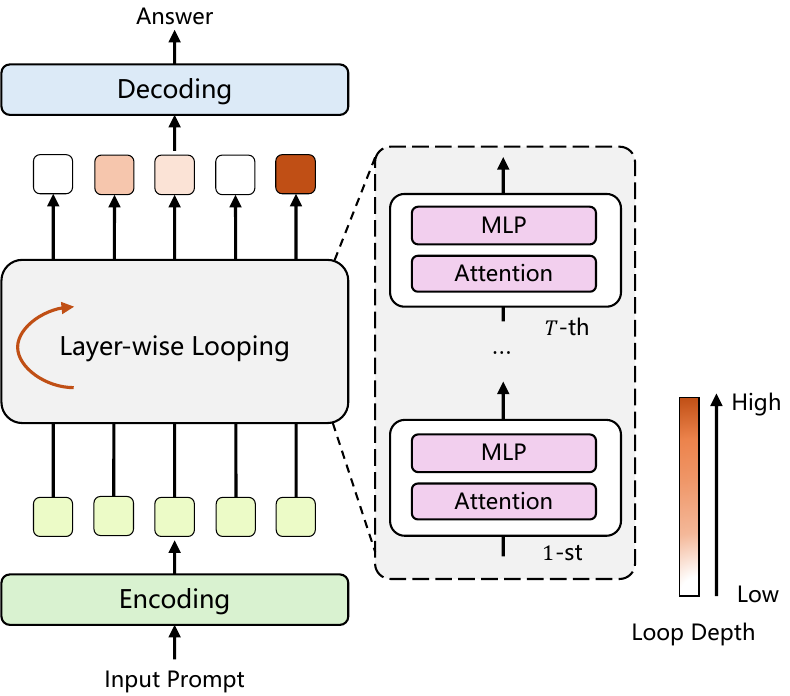}
    \caption{Illustration of layer-wise vertical-level latent CoT reasoning.}
    \label{fig:structural_cot}
    \vspace{-12pt}
\end{figure}
(\romannumeral1) \textbf{End-to-end adaptive-depth looping}, where the looped decoder is trained end-to-end so that each token dynamically decides how many internal "thinking" steps to take. The central insight is that not all reasoning steps are equally difficult; thus, the model should be able to think longer for complex tokens and proceed quickly for simpler ones. This is achieved through various architectural designs, such as implementing recurrent structures within the decoder~\citep{scaling_up_test_time}, using lightweight routers to control recursion depth on a per-token basis~\citep{inner_thinking_transformer, bae2025mixtureofrecursionslearningdynamicrecursive}, or materializing each loop's output as a latent "thought token" to improve information flow~\citep{mohtashami2024cotformerchainofthoughtdrivenarchitecture}.
(\romannumeral2) \textbf{Latent-semantic supervision inside the loop} tackles the challenge that latent thinking remains opaque and usually unsupervised, lacking explicit signals to define what constitutes a "good" latent trajectory. To address this, ~\citet{zeng2025pretraininglanguagemodelsponder} use continuous, probability-weighted embeddings to create smoother intermediate targets. ~\citet{du2025latentthinkingoptimizationlatent} employ a classifier to provide a direct ``latent reward'' that steers the reasoning trajectory.

Overall, these approaches demonstrate that reasoning capacity can be strengthened not by enlarging the parameter set but by repeatedly applying the same set of parameters along the vertical axis of the network.
By reframing depth as a flexible resource rather than a fixed architectural choice, this line of work establishes a principled "think-then-speak" paradigm that bridges the gap between fully verbalized and silent latent CoT reasoning.

\section{Analysis}
\label{sec:analysis}

To understand the internal mechanism of latent CoT, researcher explore multiple directions to examine how latent CoT models behave externally and how their internal computations give rise to reasoning. 
We organize existing analytical work into two complementary subsections: \textbf{Behavioral Observations}, which analyzes the output behavior and scaling patterns of models that employ latent CoT; and \textbf{Internal Mechanisms}, which explores how latent CoT arises within the model.

\subsection{Behavioral Observations}
\label{sec:ext-behavior}
This subsection reviews empirical studies focusing on the observable behaviors of latent CoT models, including task accuracy and scaling properties, without delving into their hidden states.
\citet{zhu20254thdimensionscalingmodel}, \citet{li2025skiplayerloopit} and \citet{saunshi2025looped} showed that repeatedly applying the same set of transformer layers—effectively increasing the model’s computational depth—markedly improved multi-step reasoning, even when the total number of parameters, and thus the amount of stored knowledge, was kept nearly constant.  
This line of evidence highlighted that \textit{reasoning ability could be strengthened by deeper computation rather than by enlarging the model size}, thereby providing empirical support for the approach discussed in \S\ref{sec:Internal Mechanism}.  
To investigate whether reasoning steps could be carried out internally rather than explicitly verbalized, \citet{liu2024languagemodelslearnskip} demonstrated that models could internalize redundant intermediate steps when producing CoT, and \citet{lin2025implicit} found that shortcut-like implicit reasoning tended to emerge when models were trained on highly regular patterns.  
These findings suggested that \textit{latent CoT could unfold entirely within hidden representations without emitting a full sequence of reasoning tokens}, providing empirical support for efforts toward internalized CoT~\citep{deng2024explicit}.  
By examining such behaviors, these analyses offered valuable insights that could inform the further development of latent-CoT approaches.

\subsection{Internal Mechanisms}
\label{sec:Internal_Mechanisms}
Whereas behaviors describe how latent CoT manifests at the output level, uncovering its \textit{internal mechanisms} requires examining the model’s representational dynamics and its theoretical reasoning capacity.
We group existing analyses into two complementary strands: \emph{Theoretical analysis}, which compares the reasoning capacity of different latent CoT paradigms against explicit CoT; and \emph{empirical analysis}, which probes hidden-state to study how latent CoT unfolds during inference.

\paragraph{Theoretical Analysis}
Recent work~\citep{zhu2025reasoningsuperpositiontheoreticalperspective, aime2024, zhu2025emergencesuperpositionunveilingtraining} analyzed the paradigm introduced in Section~\ref{sec:Token-wise Strategies} and showed that the \textit{Token-wise Horizontal Level} approach can solve certain structured reasoning problems with significantly fewer steps than explicit CoT.
Further, \citet{xu2025formalcomparisonchainofthoughtlatent} and \citet{saunshi2025looped} formally analyzed looped transformers (\S\ref{sec:Internal Mechanism}), showing that the \textit{Layer-wise Vertical Level} approach not only remains effective across a wide range of reasoning tasks but also supports a degree of parallel computation, making it more efficient than the inherently sequential nature of explicit CoT.
These theoretical results provide evidence that latent CoT can be more computationally efficient and theoretically advantageous than explicit CoT.

\paragraph{Empirical Analysis}
This line of work primarily probes hidden states using the \textit{logit lens}~\citep{logit_lens} to investigate whether models engage in latent CoT during inference.
Several studies reported evidence that LLMs exhibit such internalized reasoning \citep{lindsey2025biology, 
hou-etal-2023-towards, brinkmann2024mechanisticanalysistransformertrained,
yang2024largelanguagemodelslatently, shalev2024distributionalreasoningllmsparallel, zhang2025uncoveringlatentchainthought, wang2025latentspacechainofembeddingenables, yang2025internalcot}.
However, other work challenged this view: \citet{wang2024grokkedtransformersimplicitreasoners} argued that Transformers developed implicit reasoning only after extended "grokking"-style training far beyond standard overfitting; \citet{kudo2025thinktotalktalktothinkllmscome} found that models had not derived an answer upon first reading the problem but obtained (sub)answers progressively while generating the reasoning chain; and \citet{yu2025llmsreallythinkstepbystep} showed that prompting alone did not elicit latent CoT, which only emerged after training.
Further limits were identified by \citet{lin2025implicit}, who observed that latent-CoT behavior emerged only under fixed-pattern training data; \citet{csordás2025languagemodelsusedepth}, who found no systematic progression toward higher-level reasoning in deeper layers; \citet{wu2025llmssinglethreadedreasonersdemystifying}, who showed that a single token path could not branch into multiple reasoning threads.
These studies reveal an ongoing debate: some analyses detect signs of latent CoT within hidden representations, whereas others suggest such reasoning is fragile, highly conditional, or absent in many practical settings.
\section{Applications}
\label{sec:application}

Latent CoT reasoning has been successfully applied in natural language-based textual reasoning tasks (refer to Appendix~\ref{sec:textreasoning}). Below, we discuss other representative applications.

\paragraph{Multimodal Reasoning and Generation.}  
Latent CoT has recently been extended to multimodal domains, where generating step-by-step explanations in natural language becomes both inefficient and semantically brittle. 
To address this, recent visual approaches~\citep{shen2025efficientreasoninghiddenthinking, liu2025ssrenhancingdepthperception, li2025latentvisualreasoning, yang2025machinementalimageryempower, pham2025multimodalchaincontinuousthought, chen2025reasoningdarkinterleavedvisiontext, sun2025latentchainofthoughtvisualreasoning} embed intermediate reasoning within latent visual or spatial representations, enhancing depth perception and fine-grained understanding. 
Building on these capabilities, latent world models in autonomous driving leverage this paradigm to predict future states and optimize planning by reasoning over action-aligned latent representations~\citep{li2025enhancing, tan2025latentchainofthoughtworldmodeling}. 
The versatility of Latent CoT extends to the auditory domain, such as speech synthesis~\citep{xue2025enhancing}, and broader unified multimodal generation~\citep{sun2024multimodall}. 
As modalities diversify, the ability to steer these latent spaces may become the key to controllable, efficient multimodal intelligence.

\paragraph{Information Retrieval and Recommendation.} 
Latent CoT reasoning has been applied to enhance information retrieval systems~\citep{zhang2025reinforcedlatentreasoningllmbased}, where conventional single-pass encoding methods often fail to capture the complex semantics of documents or the evolving nature of user preferences. In dense retrieval, this involves creating richer document embeddings through a chain of deliberation~\citep{ji2025learningeffectiverepresentationsdense}. 
In recommendation systems, it allows for a deeper understanding of user interests by recurrently refining the user's interaction history in latent spaces~\citep{tang2025thinkrecommendunleashinglatent, liu2025lareslatentreasoningsequential} or by reasoning over noisy cross-domain signals, such as search histories~\citep{shi2025bridgingsearchrecommendationlatent}. 
\section{Challenges and Future Directions}
\label{sec:challenges}

\paragraph{Challenges}
(1) \textbf{Training Instabilities.}
Current methods still underperform explicit CoT approaches, largely due to the instability of training. Developing effective latent training methods remains a key challenge.
(2) \textbf{Generalization Difficulties.}
Models trained with latent CoT techniques often struggle with novel problem structures or reasoning patterns not encountered during training~\citep{lin2025implicitreasoningtransformersreasoning}. 
This fragility underscores the considerable challenge that remains in developing truly generalizable reasoning capabilities within latent spaces.
(3) \textbf{Interpretability Concerns.}
Recent studies indicate that models often perform reasoning in their ``heads'' that is not reflected in their verbalized CoTs, raising concerns about unfaithful or hidden internal processes~\citep{chen2025reasoning, lindsey2025biology}. This shift to latent CoT introduces the challenge for understanding how models draw a particular conclusion.

\paragraph{Future Directions}
\label{subsec:Future}
(1) \textbf{Alternative Architectures.} Beyond conventional Transformer architectures, recurrent Transformer variants, such as pre-trained Looped~\citep{zhu2025scalinglatentreasoninglooped}, Recursive~\citep{zhang2025recursivelanguagemodels} and Diffusion Language Models~\citep{ye2024diffusionthoughtschainofthoughtreasoning, huang2025reinforcingdiffusionchainlateral, kang2025ladirlatentdiffusionenhances}.  
(2) \textbf{Interpretability, Verification, and Privacy.} Developing effective methods to probe, interpret, or verify latent representations is crucial for improving transparency and calibrating latent CoT reasoning behavior~\citep{chen2025sealsteerablereasoningcalibration}. Leveraging this implicit nature to shield sensitive data and proprietary logic from prompt leakage presents a promising avenue for privacy-preserving reasoning~\citep{green-etal-2025-leaky}.
(3) \textbf{Training Approaches.} Reinforcement learning provides a viable solution for exploring the potential of LLMs to enhance latent CoT~\citep{guo2025deepseek}. Curriculum learning that adopts a simple-to-complex training process is also promising.
(4) \textbf{Autonomous Agents and Multi-Agent Systems.} Conventional agents often generate lengthy and verbose reasoning sequences, introducing substantial computational overhead~\citep{zhou2025largereasoning, li2024personalllmagentsinsights, zhang2024efficientllmgroundingembodied}. Latent CoT reasoning addresses this by enabling more compact decision-making and faster planning for agents. In multi-agent systems, a shared latent working memory enables lossless information exchange and direct, efficient collaboration~\citep{zou2025latentcollaborationmultiagentsystems}.
(5) \textbf{Social Intelligence.} Latent CoT provides a natural substrate for modeling nested mental states essential to \textit{Theory of Mind}~\citep{ma-etal-2023-towards-holistic}. Embedding latent belief modeling into reasoning pipelines could offer a scalable path toward socially competent AI.
\section{Conclusion}
\label{sec:conclusion}
\vspace{-5pt}
This paper presents a comprehensive survey of latent CoT reasoning in LLMs. Moving reasoning beyond surface-level language into the latent space enables more abstract, efficient, and scalable inference. We summarize the key methods, identify major challenges, and highlight promising future directions. The promise of latent CoT extends beyond optimizing inference; it points toward a future where models can reason in ways not strictly confined to language. We hope this survey lays a solid foundation and offers valuable insights to support further exploration in this emerging field.

% \newpage
\section*{Limitations}
\label{sec:limitation}

While this survey offers a comprehensive review of existing methodologies and analyses of latent Chain-of-Thought (CoT) reasoning with LLMs, a few limitations remain in the areas of interpretability, internal analysis, and alignment. Due to the breadth and rapid evolution of this emerging field, we may have inadvertently omitted other valuable contributions. We highlight several promising future directions, including alternative architectures, training paradigms, and LLM agents. Additionally, as many surveyed works rely on small-scale models or limited benchmarks, there is a need for more up-to-date and rigorous empirical validation. We advocate for continued, in-depth research to provide practitioners with actionable and robust insights into the design and deployment of latent reasoning models.

\section*{Ethics Statement}
\label{subsec:ethics} 
This survey is based entirely on publicly available research papers, models, and datasets. All referenced works are properly cited and used in accordance with their respective licenses and intended purposes. While latent reasoning introduces some intrinsic challenges, this survey aims to provide a neutral, structured overview of the field without promoting specific deployments. We emphasize the importance of future work addressing interpretability, safety, and transparency in latent reasoning.

\section*{Acknowledgements}
\label{subsec:ack} 
We thank all anonymous reviewers for their valuable comments during the review process. This work was supported by Research Grants Council of Hong Kong (GRF No. 15209724).

% Bibliography entries for the entire Anthology, followed by custom entries
%\bibliography{anthology,custom}
% Custom bibliography entries only
\bibliography{custom}

\newpage
\appendix
\section*{Appendix}
\label{sec:appendix}
\section{Pioneer Work on Latent CoT}
\label{sec:pioneerwork}
\subsection{Discrete Tokens}
\label{sec:discrete tokens}

Discrete tokens, which serve as symbolic representations of intermediate reasoning steps or cognitive operations, have emerged as a promising paradigm for enhancing the reasoning capabilities of LLMs. They significantly contribute to improved task performance and greater efficiency.

Early studies in exploring discrete tokens introduced simple markers such as “[pause]” or ellipses (“...”) to segment reasoning steps, which has significantly improved multi-step task performance (\citet{pfau2024let}, \citet{herel2024thinking}). 
Prior to these efforts, \citet{goyal2023think} proposed adaptive and learnable “pause tokens,” which dynamically allocate computational resources. These tokens enable delayed prediction, allowing models to perform additional internal computation before generating outputs, thereby enhancing accuracy for logic-intensive tasks.
\citet{pfau2024let} pointed out that the structural organization of tokens is more critical than their semantic content. Surprisingly, replacing meaningful tokens with neutral placeholders yields negligible performance loss, underscoring the importance of token structure. Subsequent analysis provided a mechanistic explanation for this phenomenon, revealing that the added computation induces a beneficial redistribution of activations within the model's internal layers, thereby enhancing its representational capacity~\citep{shi2025meaninglesstokensmeaningfulgains}.

Beyond these pioneering exploration, researchers developed more sophisticated tokens to encode complex reasoning structures. For example, \citet{wang2023guiding} introduced ``planning tokens'' derived from heuristics or variational autoencoders (VAEs) to improve coherence and precision in reasoning. To disentangle cognitive processes and enhance interpretability, \citet{jin2024disentangling} proposed specialized tokens such as ``memory'' and ``reason'', which modularize reasoning by isolating specific cognitive operations.

Overall, discrete tokens have progressed from simple markers to versatile tools for abstract cognitive modeling. They serve as powerful mechanisms that advances LLM reasoning capabilities, improving both efficiency and interpretability.

\subsection{Implicit CoT}
\label{sec:implicitcot}

In addition to the exploration of meaningless token-driven reasoning, another promising avenue involves internalizing explicit CoT directly into the latent representations of LLMs.
ICoT \citep{deng2023distillcot} aligns hidden states between a teacher generating CoT and a student producing direct answers, enabling implicit reasoning through intermediate representation supervision.
System 2 Distillation \citep{weston2024system2} introduces a self-supervised method to compile various multi-step reasoning strategies into a standard model, distilling the benefits of System 2 thinking into faster, token-efficient System 1 outputs.
Stepwise Internalization \citep{deng2024explicit} proposes a staged fine-tuning procedure that gradually removes intermediate reasoning supervision, allowing models to internalize reasoning while improving robustness and efficiency.
TwT \citep{xu2025twtthinkingtokenshabitual} presents a three-stage habitual reasoning distillation framework using multi-teacher guidance and dual-criteria sampling, compressing explicit reasoning into latent capabilities with minimal inference overhead.
Overall, this line of work shows that \textit{it is effective to condense reasoning processes into compact and computationally efficient latent structures.}

\section{Another Emerging Method for Representation Initialization}
\label{sec:special_vector}
\paragraph{Special Vector}  This type of methods initialize latent representation using special vector parameters specifically for reasoning. One strategy is to augment the model with an internal, learnable scratchpad. This is implemented as a fixed sequence of placeholder tokens~\citep{jiang2025dartdistillingautoregressivereasoning, wang2025synadaptlearningadaptivereasoning}, a set of new ``latent tokens'' inserted directly into the input~\citep{sun2025enhancinglatentcomputationtransformers}, or a more complex learnable state that serves as the starting point for reasoning~\citep{liu2025marcosdeepthinkingmarkov, zheng2025fastthinkinglargelanguage, zhang2025lightthinkerthinkingstepbystepcompression}. 
A different strategy offloads this task to an auxiliary model, such as quantizing text into new latent tokens using a VQ-VAE~\citep{su2025token} or extracting concept vectors via a sparse autoencoder~\citep{tack2025cocomix}.

\section{Common Tasks of Textual Reasoning}
\label{sec:textreasoning}
General tasks of latent CoT reasoning include mathematical reasoning~\citep{cobbe2021trainingverifierssolvemath, deng2023distillcot, hendrycks2021measuring, miao2021diversecorpusevaluatingdeveloping, patel2021nlpmodelsreallyable, ling2017programinductionrationalegeneration}, general commonsense reasoning~\citep{talmor-etal-2019-commonsenseqa, suzgun-etal-2023-challenging, rein2023gpqagraduatelevelgoogleproofqa, hendrycks2021measuringmassivemultitasklanguage}, and logical multi-hop reasoning datasets~\citep{yang2018hotpotqadatasetdiverseexplainable, geva2021didaristotleuselaptop, saparov2023languagemodelsgreedyreasoners,  hao2024traininglargelanguagemodels}. Recent latent reasoning methods also evaluate on several high-bar reasoning benchmarks that have become standard for assessing Large Reasoning Models~\citep{aime2024}, and code-centric datasets~\citep{jimenez2024swebenchlanguagemodelsresolve, jain2024livecodebenchholisticcontaminationfree}. Moreover, there remains a lack of benchmarks that are both aligned with real-world applications and specifically designed to showcase the advantages of latent CoT reasoning.

\begin{table*}[htbp]
\centering
\small 
\renewcommand{\arraystretch}{1.2} 
% 使用 @{} 去除最左和最右的默认空白，微调 p{} 的宽度以完美适配 16cm 的 textwidth
\begin{tabular}{@{} >{\raggedright\arraybackslash}p{1.6cm} >{\raggedright\arraybackslash}p{2.2cm} >{\raggedright\arraybackslash}p{2.8cm} >{\raggedright\arraybackslash}p{2.1cm} >{\raggedright\arraybackslash}p{2.5cm} >{\raggedright\arraybackslash}p{2.8cm} @{}}
\toprule
\textbf{Representation Category} & \textbf{Method} & \textbf{Optimization Objective} & \textbf{Model} & \textbf{Training Data} & \textbf{Test Data} \\
\midrule
\textbf{---} 
 & No-CoT-SFT & SFT: $Q \rightarrow A$ & GPT-2 & GSM8K-Aug & GSM8K \\
 & CoT-SFT & SFT: $Q \rightarrow CoT \rightarrow A$ & GPT-2 & GSM8K-Aug & GSM8K \\
\midrule
\textbf{Hidden State} 
 & Coconut~\citep{hao2024traininglargelanguagemodels} & Multi-stage curriculum (Language to Continuous) & GPT-2 & GSM8K-Aug, ProntoQA, ProsQA & GSM8K, ProntoQA, ProsQA \\
 & CODI~\citep{shen2025codi} & Self-distillation + Feature alignment & GPT-2, LLaMA 3.2 1B-it & GSM8K-Aug, GSM8K-Aug-NL & GSM8K, SVAMP, GSM-Hard, MultiArith \\
 & PCCoT~\citep{wu2025parallelcontinuouschainofthoughtjacobi} & Self-distillation + Jacobi iteration & GPT-2, LLaMA 3.2 1B-it & GSM8K-Aug, GSM8K-Aug-NL & GSM8K \\
 & SemCoT~\citep{he2025semcotacceleratingchainofthoughtreasoning} & Contrastive alignment + Answer CE loss & Llama-2-7B, Mistral-7B & GSM8K, SVAMP, MultiArith, CSQA, CoinFlip & GSM8K, SVAMP, MultiArith, CSQA, CoinFlip \\
 & SIM-CoT~\citep{wei2025simcotsupervisedimplicitchainofthought} & Step-level reconstruction + Answer CE loss & GPT-2, LLaMA 3 (1B/3B/8B) & GSM8K-Aug, GSM8K-Aug-NL & GSM8K, GSM-Hard, SVAMP, MultiArith \\
 & System-1.5 Reasoning~\citep{wang2025system15reasoningtraversallanguage} & Alignment + Shortcut learning (Router-adapters) & GPT-2, LLaMA 3.2 1B & GSM8K-Aug, StrategyQA & GSM8K, StrategyQA, GSM-Hard \\
 & KaVa~\citep{kuzina2025kavalatentreasoningcompressed} & Self-distillation + KV-cache matching & LLaMA 3 (1B/3B/8B) & GSM8K-Aug, GSM8K-Aug-NL & GSM8K, GSM-Hard, SVAMP, MultiArith \\
\midrule
\textbf{Weighted Embedding} 
 & CoLaR~\citep{tan2025thinksilentlythinkfast} & Next-embedding prediction + RL (GRPO) & LLaMA 3.2 1B-it & GSM8K-Aug & GSM8K, GSM-Hard, SVAMP, MultiArith \\
 & HRPO~\citep{yue2025hybridlatentreasoningreinforcement} & On-policy RL (Outcome reward + KL) & Qwen2.5 (1.5B/3B)-it & NQ, HotpotQA, MMLU, ARC-C, GSM8K, MATH & NQ, TriviaQA, HotpotQA, 2WikiMQA, Bamboogle, GSM8K, MATH, MMLU-ST, ARC-C \\
 & Soft Thinking~\citep{zhang2025softthinkingunlockingreasoning} & Training-free (Inference only) & QwQ-32B, DeepSeek-R1-Distill-Qwen-32B, DeepSeek-R1-Distill-Llama-70B & --- & GSM8K, Math500, AIME 24, GPQA, HumanEval, MBPP, LCB \\
\midrule
\textbf{Special Vector} 
 & LightThinker~\citep{zhang2025lightthinkerthinkingstepbystepcompression} & SFT + Thought-based attention masking & Qwen2.5-7B, Llama3.1-8B & Bespoke-Stratos-17k & GSM8K, MMLU, GPQA, BBH \\
\bottomrule
\end{tabular}
\caption{A structured evaluation framework detailing the technical implementations, optimization objectives, and dataset configurations of representative latent reasoning approaches. Results and settings are sourced from their respective original papers.}
\label{tab:technical_comparison}
\end{table*}

\end{document}